\documentclass[runningheads]{llncs}

\usepackage{eccv}

\usepackage{eccvabbrv}

\usepackage{graphicx}
\usepackage{booktabs}

\usepackage[accsupp]{axessibility}  % Improves PDF readability for those with disabilities.

\usepackage{hyperref}

\usepackage{orcidlink}

\begin{document}

\title{GarmentAligner: Text-to-Garment Generation via Retrieval-augmented Multi-level Corrections} 

\titlerunning{GarmentAligner}

\author{
Shiyue Zhang\inst{1} \and
Zheng Chong\inst{1,3} \and
Xujie Zhang\inst{1} \and
Hanhui Li\inst{1} \and
Yuhao Cheng\inst{2} \and
Yiqiang Yan\inst{2} \and
% yiqiang yan\inst{3}\orcidlink{2222--3333-4444-5555} \and
Xiaodan Liang\inst{1,3}\thanks{Corresponding author.}}
% \thanks{Corresponding author.}
\authorrunning{S.~Zhang et al.}

\institute{
Shenzhen Campus of Sun Yat-sen University, Shenzhen, China \and
Lenovo Research, Shenzhen, China \and
Research Institute of Multiple Agents and Embodied Intelligence, Peng Cheng Laboratory, Shenzhen, China\\ 
\email{xdliang328@gmail.com}
}
\maketitle

{
\centering
\captionsetup{type=figure}
\includegraphics[scale=0.36]{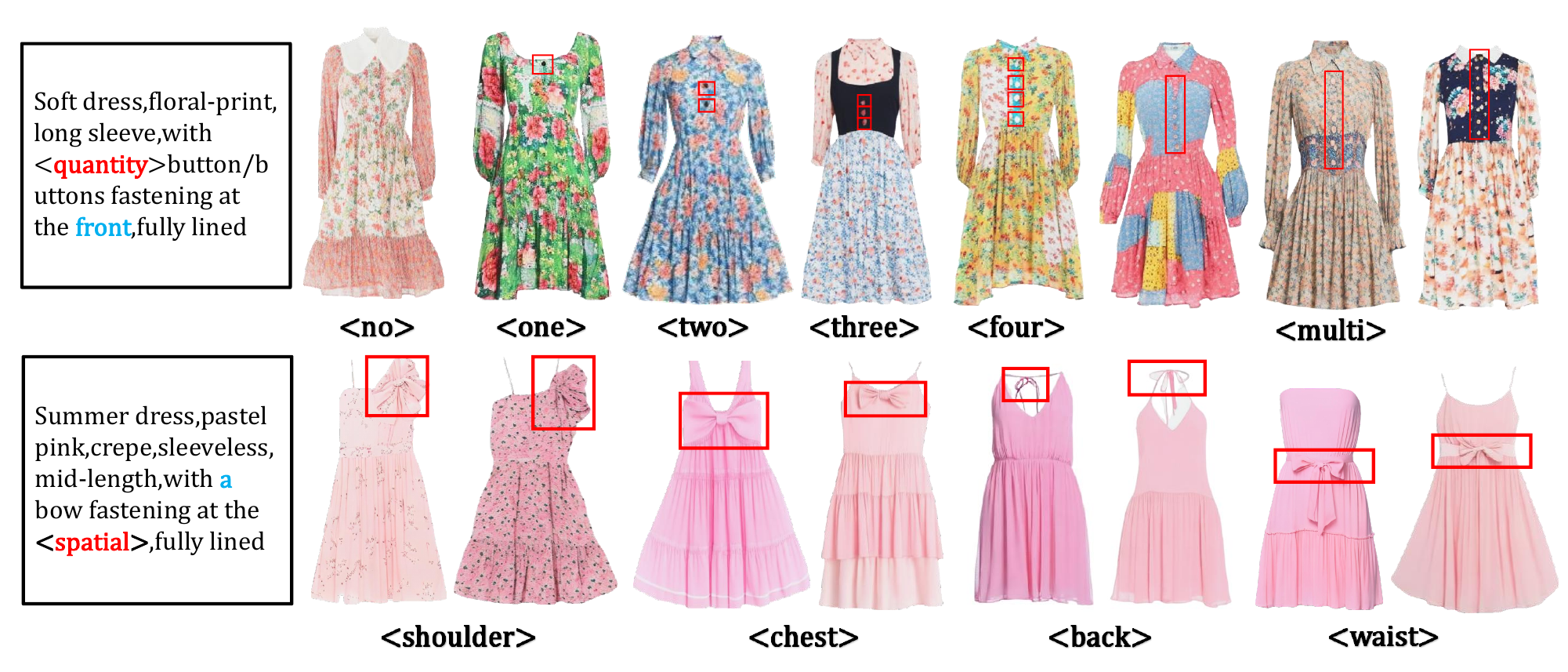}
\captionof{figure}{GarmentAligner is capable of producing high-quality garment images accurately depicting the quantity and spatial alignment of components specified in the provided captions.}
\label{fig:teaser}
\vspace{-6mm}
}

\begin{abstract}
General text-to-image models bring revolutionary innovation to the fields of arts, design, and media. 
However, when applied to garment generation, even the state-of-the-art text-to-image models suffer from fine-grained semantic misalignment, particularly concerning the quantity, position, and interrelations of garment components.
Addressing this, we propose \textit{GarmentAligner}, a text-to-garment diffusion model trained with retrieval-augmented multi-level corrections. 
To achieve semantic alignment at the component level, we introduce an automatic component extraction pipeline to obtain spatial and quantitative information of garment components from corresponding images and captions.
% design a data preprocessing pipeline to automatically extract spatial and quantitative information about garment components from corresponding images and captions.
% \hh{the ``data preprocessing pipeline" term sounds boring. Maybe like: we first design an automatic component extraction pipeline to obtain spatial and quantitative information of garment components from corresponding images and captions.}
Subsequently, to exploit component relationships within the garment images, we construct retrieval subsets for each garment by retrieval augmentation based on component-level similarity ranking and conduct contrastive learning to enhance the model perception of components from positive and negative samples.
% expanding the available training data to a 10\(\times\) scale.
% \hh{perhaps, here we need to emphasize the racl method can find related components to tackle an issue, instead of just data scaling. Because our total dataset size is about 1M (not limited), and conventional data augmentations like random cropping can also generate 10x data easily.}
To further enhance the alignment of components across semantic, spatial, and quantitative granularities, we propose the utilization of multi-level correction losses that leverage detailed component information.
The experimental findings demonstrate that GarmentAligner achieves superior fidelity and fine-grained semantic alignment when compared to existing competitors.

\end{abstract}

\section{Introduction}
\label{sec:intro}

\begin{figure}[!t]
  \centering
  \includegraphics[scale=0.32]{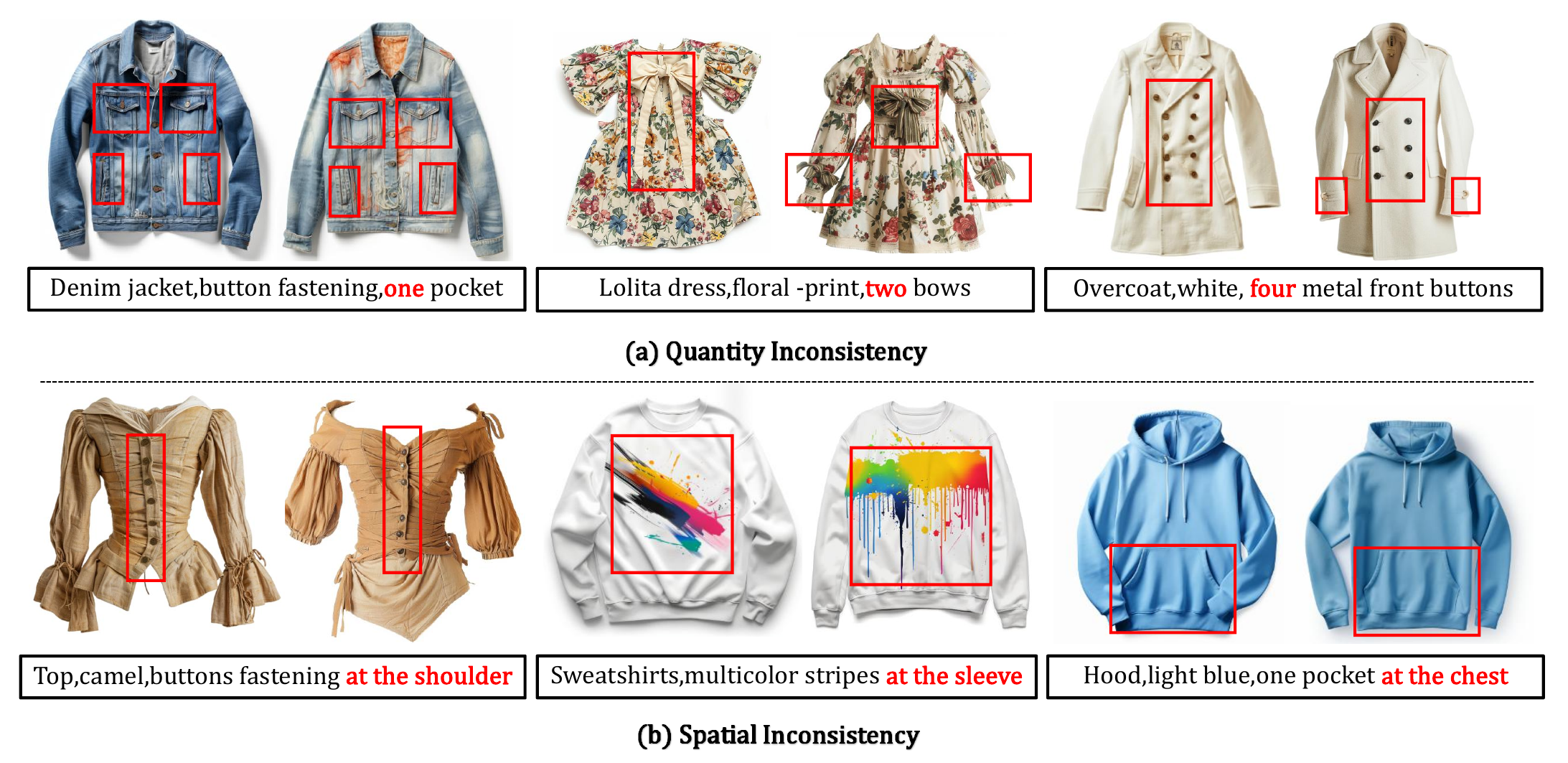}
  \caption{The failures of state-of-the-art text-to-image model Midjourney in the text-to-garment task. The misalignment is primarily attributed to the quantities and spatial positions of components, thereby making it difficult to generate garments that meet the expected fine-grained details.} 
   \label{fig.inconsi}
  \vspace{-3mm}
\end{figure}

Text-to-garment generation, as one of the downstream tasks of text-to-image generation \cite{razzhigaev2023kandinsky, podell2023sdxl, rombach2022highresolution}, has emerged as a promising technology in the field of computer vision, offering the potential to revolutionize the fashion industry by providing creative inspirations and accelerating the design process.

Despite the rapid developments in general text-to-image models\cite{razzhigaev2023kandinsky, podell2023sdxl, rombach2022highresolution}, adapting these advanced models for garment generation tasks presents significant challenges.
% \xd{SOLVED: here should have one figure to illustrate the limitation of existing sd-xl or MJ, highlight their drawbacks in detail with highlighted boxes and our advantages}
On the one hand, the visual semantics in the fashion domain significantly differ from those in general text-to-image generation. Garment captions possess specific textual structures and professional modifiers. Existing text-to-image models\cite{razzhigaev2023kandinsky, podell2023sdxl, rombach2022highresolution} struggle to achieve high consistency between textual descriptions and the generated garment images, as illustrated in \cref{fig.inconsi}.
While some methods for garment generation\cite{Zhang_2022armani, zhang2023diffcloth, seyfioglu2023dreampaint, xie2024hierarchical, he2024dresscode} to capture the semantic structure of garment captions, thereby partially alleviating textual disparities, achieving fine-grained alignment still remains challenging.
On the other hand, garment components, as crucial elements within garments, encompass distinct attributes and intricate interconnections, posing a challenge in the accurate generation of authentic garment images that satisfy the diverse requirements of these components. While several existing methods \cite{xie2024hierarchical, Zhang_2022armani, zhang2023diffcloth, yu2023quality} emphasize the importance of garment components, they primarily focus on the visual semantic aspect, neglecting the exploration of more extensive details such as component positioning and quantity.

\begin{figure}[!t]
    \label{fig:retrievalMoti}
    \centering
    \includegraphics[scale=0.36]{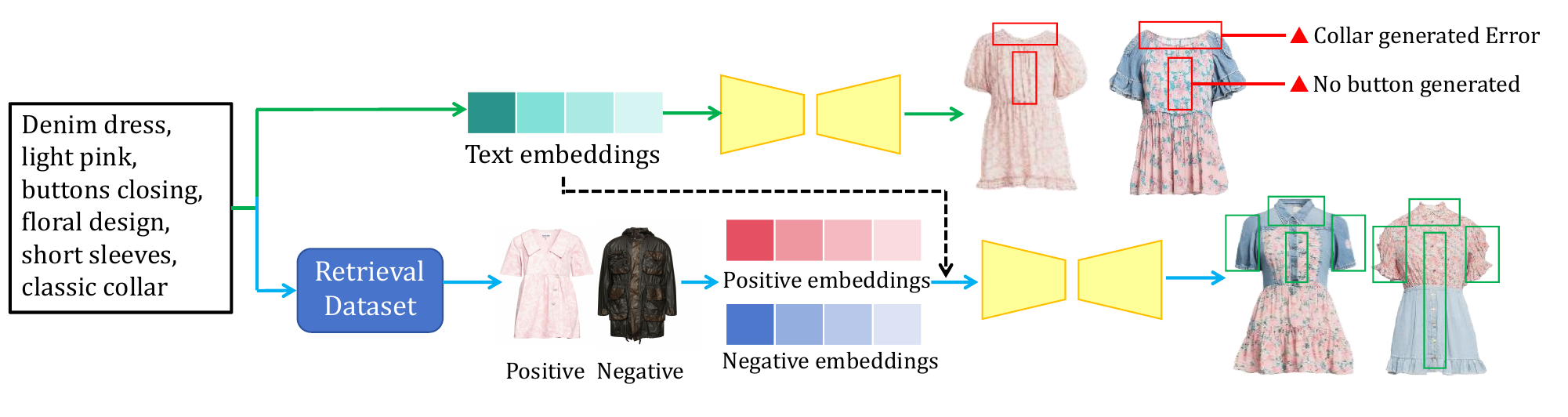}
    \caption{The illustration of misalignment is addressed by retrieval-augmented contrastive learning. By assimilating insights from positive and negative samples retrieved via component-level similarity ranking, the model enhances its perception of component relationships.} 
    \label{fig:intro_retrieval_aug}
    \vspace{-3mm}
\end{figure}

Our key insight is to explore semantic information at various levels within garment images and their corresponding captions to enhance the overall quality and fine-grained details of generated garment images. 
To realize this objective, we initially devise an automated component extraction pipeline aimed at retrieving spatial and quantitative information of garment components from the associated images and captions.
Nevertheless, optimizing the model directly using text-image pairs falls short of fully exploiting component-level information. This limitation confines the model to comprehend garment semantics solely at a holistic level.

To address this issue, we propose a training strategy integrating retrieval-augmented multi-level corrections to effectively adapt the general text-to-image diffusion model to the text-to-garment generation task.
To enhance garment semantics and perception of component relationships, we implement a multi-level semantic similarity ranking approach for each sample. This approach utilizes extracted component-level information to construct its retrieval subset. Subsequently, pairs of positive and negative samples are extracted from the retrieval subset for contrastive learning, expanding the available training samples and learning the inherent relationship between components. The misalignment addressed by retrieval-augmented contrastive
learning is shown in \cref{fig:intro_retrieval_aug}.
Additionally, we propose corrections from three perspectives: visual perception, spatial alignment, and component quantity. Specifically, we utilize the CLIP Score \cite{hessel2022clipscore} as the reward function for visual corrections, align spatial positioning using attention maps derived from text-conditioned cross-attention coupled with component segmentation, and ensure the generated components match the quantity specified in captions by conducting component detection during the training phase.
% \xd{do not need to say the whole framework, the introduction should try to attract reviewers using examples and motivation figure} The overall framework of our approach is shown in \cref{fig:architechture}.
% \xd{dataset, where dataset comes from, the benefits over prior datasets?}
% \xd{which dataset are our experiments conducted? better list the dataset name. As if you say the dataset novelty, people may argue you only did experiments on your dataset, not general ones. It will be very dangerous.}
% 
    % \item [$ \bullet $] We construct a garment dataset with rich annotations\xd{not annotations except you explain where the annotations come from or are they manually labeled? otherwise it would be very misleading}, comprising XX,XXX high-quality garment images, detailed captions, component-level semantic segmentations, and component counts.\xd{move to last part, should focus on the method novelty rather than dataset novelty}
The contributions of GarmentAligner can be summarized as follows:
\begin{itemize}
% \vspace{-5mm}
    \item [$ \bullet $] GarmentAligner introduces a retrieval-augmented contrastive learning strategy for the text-to-garment generation task, fully leveraging the rich information extracted from garment components to enhance both the optimization of garment semantics and the perception of component relationships.
    
    \item [$ \bullet $] GarmentAligner employs three component-level corrections, addressing visual, spatial, and quantitative aspects, thereby enhancing the fine-grained multi-level alignment of the text-to-garment model.

    \item [$ \bullet $] Extensive comparative experiments conducted on the CM-Fashion dataset \cite{Zhang_2022armani} with existing methods\cite{ramesh2021dalle,Zhang_2022armani,rombach2022highresolution,tang2023composable,feng2023StructureDiffusion,chefer2023attend,zhang2023diffcloth,podell2023sdxl,chen2023pixart,xue2024raphael} demonstrate the superiority of GarmentAligner in fine-grained multi-level semantic alignment for garment components, and detailed ablation studies validate the effectiveness of different parts of GarmentAligner.

\end{itemize}

\section{Related Work}

\textbf{Text-guided image Generation.}
Many profound deep generative models \cite{podell2023sdxl, rombach2022highresolution, razzhigaev2023kandinsky, saharia2022imagen, ramesh2021dalle} have emerged in the field of text-to-image generation, especially large-scale auto-regressive models and diffusion models.
In the realm of autoregressive models, DALL-E\cite{ramesh2021dalle} exhibited remarkable zero-shot capability, while Parti\cite{yu2022parti} illustrated the potential for scaling up autoregressive models. 
However, diffusion models, renowned for their capacity to progressively synthesize images from random noise, are predominantly favored due to their high diversity and sample quality. 
Notably, DALL-E 2\cite{ramesh2022dalle2} operates within the image space of CLIP\cite{radford2021clip}to generate images, while Imagen\cite{saharia2022imagen} underscores the significance of leveraging pre-trained language models, employing a frozen T5 encoder \cite{raffel2023t5}.
The latent diffusion model \cite{rombach2022highresolution} advocates encoding images with an autoencoder and subsequently utilizing diffusion models to generate continuous feature maps in latent space, thereby enhancing efficiency.
Improving text-image consistency and enhancing image quality are fundamental objectives within the text-to-image generation domain. Numerous techniques have been explored for this purpose, including textual inversion\cite{gal2022TextualInversion}, attention modulation\cite{chefer2023attend}, spatial control\cite{avrahami2023spatext,voynov2023sketch}, and human feedback\cite{lee2023aligning,xu2024imagereward,wu2023hps} and so on, all yielding promising results.
Despite these advancements, persistent fine-grained semantic imperfections remain in the generated outputs, particularly regarding the accurate positioning and quantity of depicted objects.

\noindent\textbf{Garment Generation.}
As text-to-image generation emerges as one of the most impressive and promising applications in computer vision, downstream tasks in various scenarios are also rapidly evolving.
Text-to-garment generation, as one of the downstream tasks, has garnered considerable attention due to its potential commercial value in the fashion design industry. 
Some text-to-garment methods  \cite{xie2024hierarchical, ning2023picture, li2023unihuman, yu2023quality, baldrati2023multimodal, Lin_2023fashiontex} focus on the generation and manipulation of garments on human bodies, while others like \cite{Zhang_2022armani,zhang2023diffcloth,he2024dresscode} directly convert given captions into in-shop garment images. 
Most of these methods \cite{ning2023picture, baldrati2023multimodal, zhang2023diffcloth, he2024dresscode, seyfioglu2023dreampaint} adopt pre-trained text-to-image latent diffusion model \cite{rombach2022highresolution} as the backbone, adapting pre-trained models to the domain of text-to-garment through adjustments to captions or text embeddings or fine-tuning the backbone.
Some efforts have focused on the significance of clothing components. Diffcloth \cite{zhang2023diffcloth} and ARMANI \cite{Zhang_2022armani} utilize divided captions and semantic segmentation of garment images to achieve component-level alignment. HieraFashDiff \cite{xie2024hierarchical} partitions hierarchical design concepts to enable editing of components of garments worn by models. Fashion-Diffusion \cite{yu2023quality} introduces a dataset annotated at the component level.
Despite numerous explorations in the domain of text-to-garment generation, results still exhibit granularity semantic flaws, with spatial positioning and quantity alignment being particularly severe.

\noindent\textbf{Retrieval Augmentation.}
Retrieval augmentation is initially introduced to alleviate out-of-distribution performance degradation in language modeling and natural language processing. These methods \cite{lewis2021retrievalaugmented_rag, guu2020realm, borgeaud2022retro, khandelwal2020knnlm} typically employ a retriever to extract pertinent documents from an external database, subsequently employing a generator to reference these documents for prediction.
By externalizing a model's knowledge, retrieval augmentation offers advantages in accuracy, controllability, and scalability. Consequently, many methods have since incorporated this mechanism into computer vision tasks.
For instance, RAFIC\cite{lin2023rafic} utilizes retrieval augmentation to enhance few-shot image classification performance. RDMs\cite{blattmann2022semiparametric_rdms} introduce the retrieval-augmented mechanism to diffusion models by training a conditional generative model through neighbor retrieval and flexible external conditioning inference. Re-Imagen\cite{chen2022reimagen} substitutes KNN neighbors for distance in latent space, allowing retrieved neighbors to consist of both images and text. RA-CM3 \cite{yasunaga2023retrievalaugmented_racm3} proposes a unified approach to retrieve and generate any combination of text and images.
Inspired by these methodologies, we implement retrieval augmentation at multiple levels on the text-to-garment dataset to generate samples for contrastive learning, thereby guiding holistic perception enhancement within the model.

\section{Method}

\begin{figure}[!t]
  \centering
  \includegraphics[scale=0.36]{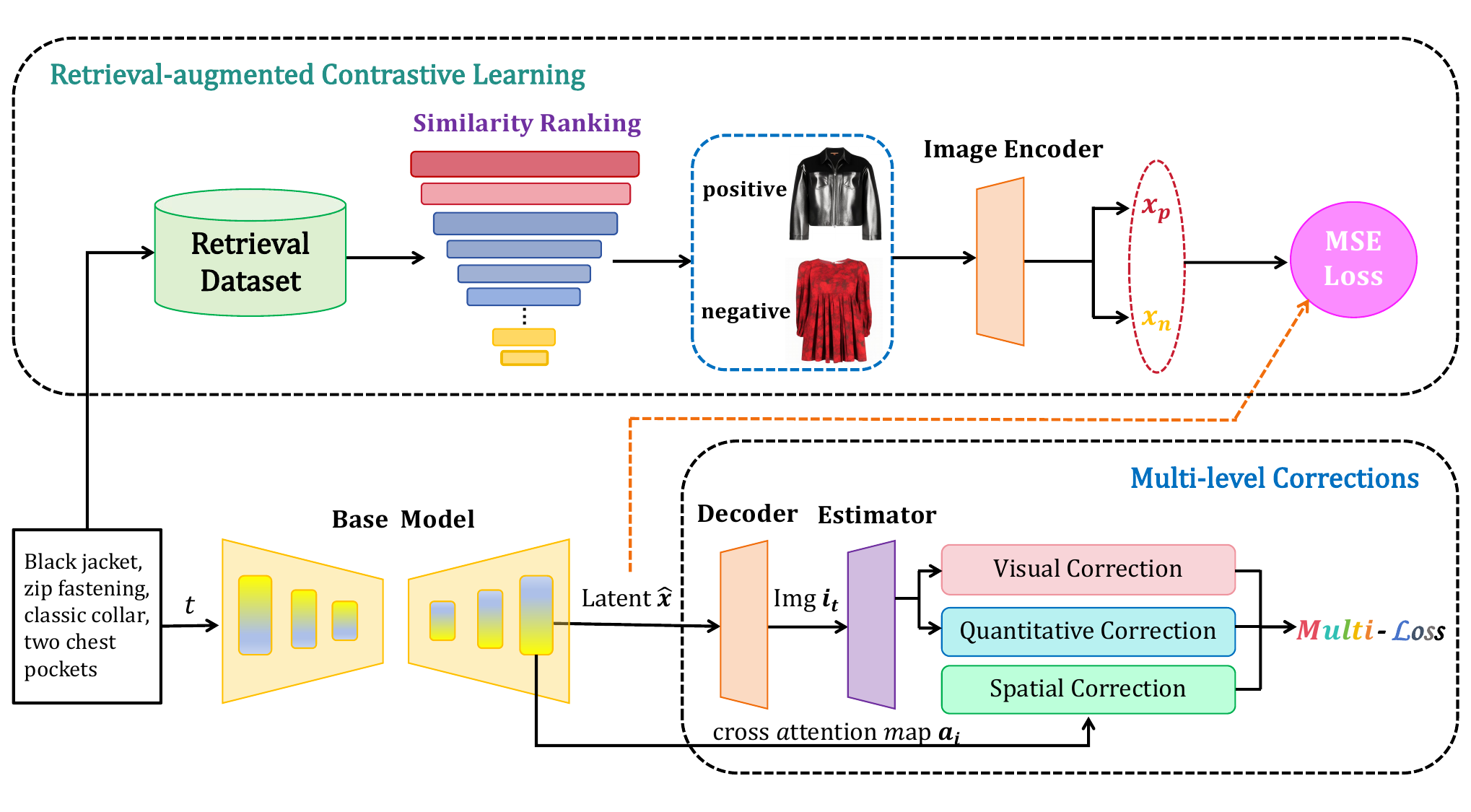}
  \caption{The overview of the proposed GarmentAligner. During training, retrieval samples are systematically constructed, utilizing multi-level semantic similarity ranking for contrastive learning, with the objective of attaining global perceptual alignment. Simultaneously, multiple correction losses are employed to refine the visual semantics, spatial positions, and the quantity of garment components, thereby augmenting the granularity of details.} 
  \label{fig:architechture}
  \vspace{-5mm}

\end{figure}

The primary objective of GarmentAligner is to enhance the alignment between input text prompts and generated garment images across multiple semantic levels in both holistic perception and fine-grained semantics.
To achieve this, GarmentAligner integrates pre-trained latent diffusion models \cite{rombach2022highresolution} as backbones to exploit their inherent knowledge and finetune the pre-trained text-to-image backbone with retrieval-augmented multi-level corrections to adapt it to the text-to-garment generation domain. Besides, an automatic component extraction pipeline (refer to \cref{sec:acep}) is adopted to attain in-depth component-level information from garment images with advanced open-domain detection and segmentation methods\cite{kirillov2023sam, liu2023groundingdino, ren2024groundedsam}.
The overview of the proposed training strategy is depicted in \cref{fig:architechture}.

The proposed GarmentAligner comprises two key parts: 

\begin{itemize}
    \item [$ \bullet $] A retrieval-augmented contrastive learning approach (refer to \cref{racl}) to utilize positive and negative samples retrieved from a subset constructed via semantic similarity ranking. This approach leverages an extensive sample pool to enhance the pre-trained model's semantic perception capabilities.
    \vspace{2mm}
    \item [$ \bullet $] Multi-level corrections (refer to \cref{mlc}) to address the deficiency in fine-grained semantic alignment of garment components across perceptual, spatial, and quantitative perspectives.
\end{itemize}

\subsection{Automatic Component Extraction Pipeline} 
\label{sec:acep}
The CM-Fashion dataset \cite{Zhang_2022armani} consists of images depicting in-shop garments accompanied by corresponding textual descriptions. However, these textual descriptions primarily emphasize overall characteristics, lacking detailed component-level quantity and spatial information. Consequently, models trained on this data often encounter misalignment issues when tasked with incorporating such specifics, thereby impeding the generation of high-quality garment images.

To attain comprehensive component-level insights, we utilize open-domain detection and segmentation models on garment images to meticulously identify and segment garment components. Specifically, we initially preprocess garment images with GroundingDINO\cite{liu2023groundingdino} to obtain bounding boxes of the targeted components. Subsequently, we determine the quantity of each component by counting these boxes, and the position of each target component is discerned by computing the geometric center of the boxes. For the segmentation at the component level, we first deploy the garment parsing model\cite{Zhang_2022armani} for primary segmentation results. We then apply SAM\cite{kirillov2023sam}, in conjunction with the garment image and the acquired boxes, to enhance the segmentation of components not previously addressed by \cite{Zhang_2022armani}. Finally, we align the information extracted from images with the textual descriptions to enrich the textual annotations with additional, precise component-level quantitative and spatial information.

Through this process, we establish a text-to-garment dataset based on the CM-Fashion dataset\cite{Zhang_2022armani} comprising garment images, detailed captions, component segmentation, position, and quantity information.

\subsection{Retrieval-augmented Contrastive Learning}
\label{racl}

\begin{figure}[!t]
  \centering
  \includegraphics[scale=0.35]{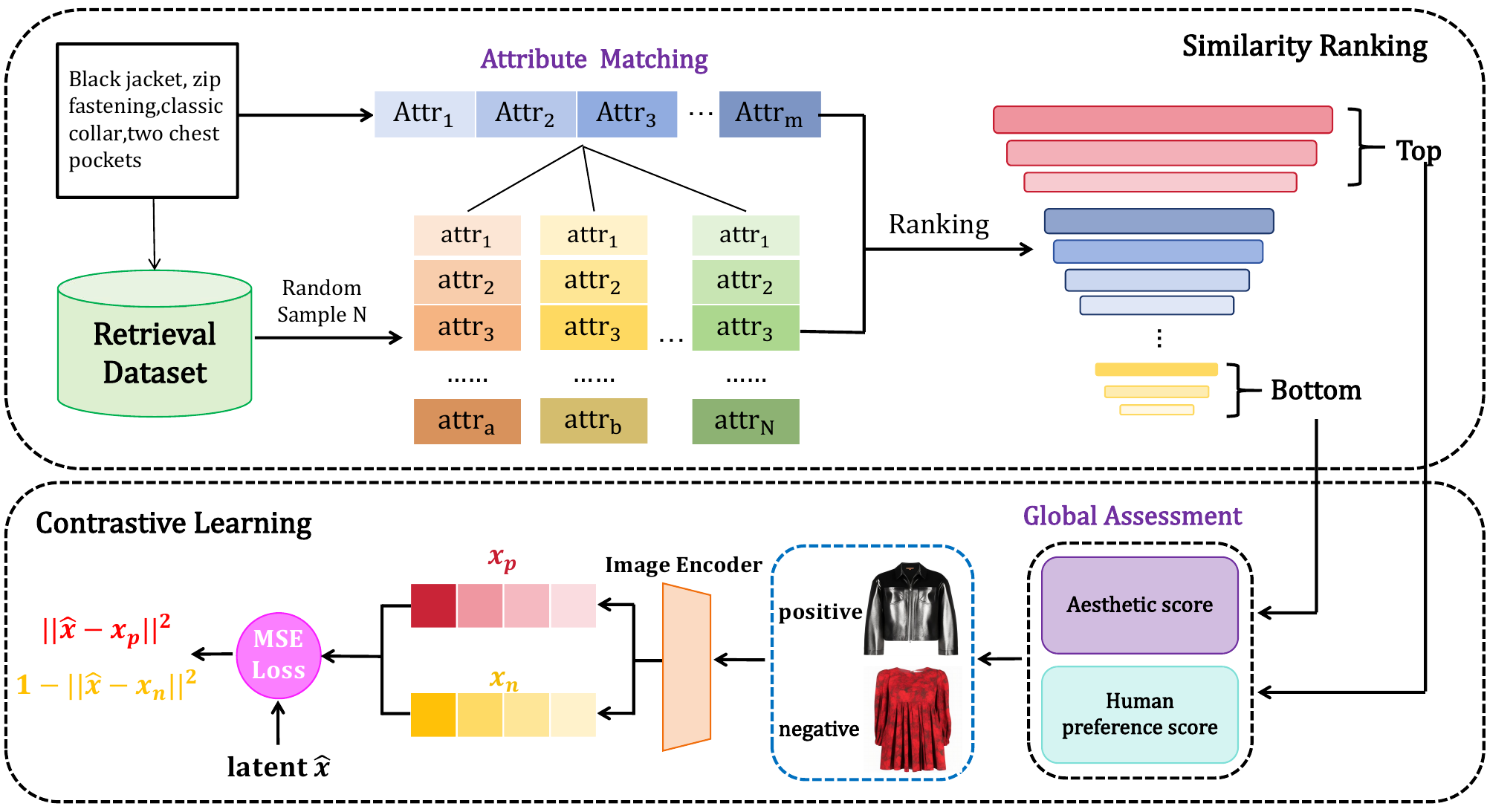}
  \caption{The illustration of proposed Retrieval-augmented Contrastive Learning. Retrieval for each sample is performed within a randomly selected subset containing \(N\) samples, based on component-level semantic similarity ranking. Subsequently, the retrieval outcomes undergo global assessment filtering to acquire positive and negative samples for contrastive learning.} 
  \label{fig:RACL}
    \vspace{-5mm}

\end{figure}

While we have constructed a high-quality text-to-garment dataset, its scale falls significantly short in comparison to general text-to-image tasks. To address the issue of data scale, we propose a retrieval-augmented contrastive learning approach for fine-tuning the text-to-garment model with even limited available data. \cref{fig:RACL} illustrates the process and details of the proposed retrieval-augmented contrastive learning.

Given a garment sample, we conduct the retrieval among a random subset with \(N\) samples by semantic similarity ranking. The semantic similarity is computed utilizing extracted information concerning component-level quantity and semantics. Specifically, for a pair of samples \((x, y)\), comprising quantities \((q^x_i, q^y_i)\) and text descriptions \((t^x_i, t^y_i)\) for the \(i\)-th component, the similarity score for the \(i\)-th component between \((x, y)\) can be expressed as:
\begin{equation}
\label{component_sim}
    S(x, y, i) = \frac{1}{| q^x_{i} - q^y_{i} | + Jaro(t^x_{i},  t^y_{i})},
\end{equation}
where \(Jaro(s_1, s_2)\) represents the Jaro distance between string  \(s_1\) and \(s_2\), and \(|\cdot|\) denotes the absolute value.
Consequently, the overall semantic similarity score of pair \((x, y)\) across all \(k\) components can be expressed as:
\begin{equation}
\label{total_sim}
    S(x, y) = \sum_{i=1}^k {S(x, y, i)} - {\alpha}\cdot Jaro(t_x,  t_y) ,
\end{equation}
where \(t_x\), \(t_y\) represent the garment captions of \(x\) and \(y\), and \(\alpha\) is the balancing weight.

The retrieval process generates a retrieval subset for each sample, enabling the expansion of a single garment sample to \(N_p \times N_n\) sample pairs for contrastive learning, based on the similarity ranking of the retrieval subset and given positive and negative sampling numbers \(N_p\) and \(N_n\). To enhance the efficacy of positive and negative sample comparisons, we incorporate aesthetic scores and human preference scores \cite{wu2023hps} for sample evaluation. This evaluation entails the selection of higher-scoring positive samples and lower-scoring negative samples, accentuating the disparity between them.

During the training phase, the optimization objective aims to minimize the Mean Squared Error (MSE) between the predicted latent result \(\hat x\) and positive samples \(x_p\), while simultaneously maximizing the MSE between \(\hat x\) and negative sample \(x_n\). Mathematically, this objective can be formulated as:
\begin{equation}
    \label{eq.3}
    {\cal L}_{\mathrm{RACL}}= ||\hat x - x_p ||^{2} +1- ||\hat x - x_n ||^{2},
\end{equation}
where \(||a - b||^2\) denotes the computation of the mean squared error between \(a\) and \(b\).

\subsection{Multi-level Corrections}
\label{mlc}
\begin{figure}[!t]
  \centering
  \includegraphics[scale=0.4]{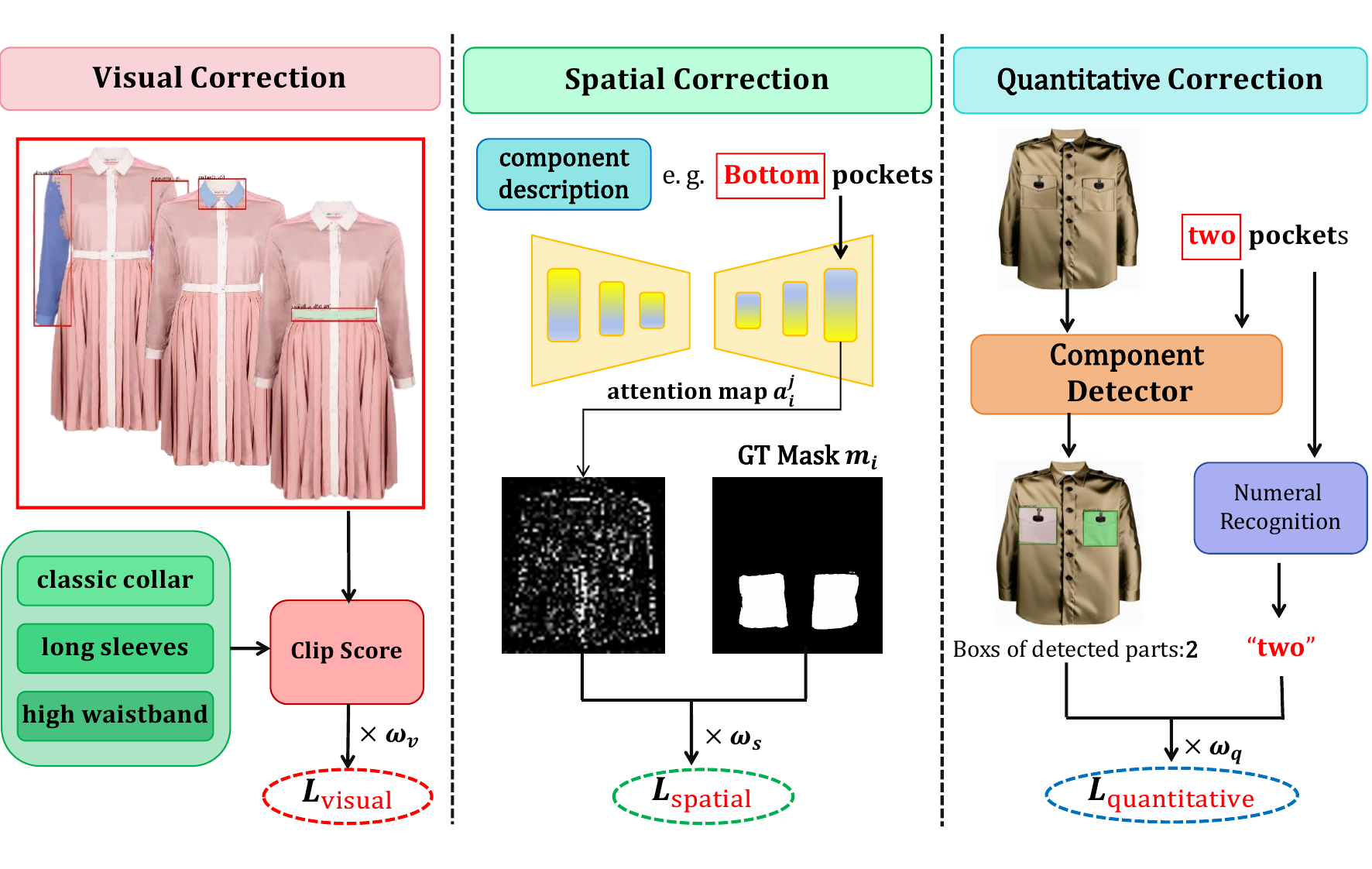}
  \caption{The illustration of the proposed Multi-level Corrections. The generated garment images undergo decomposition at the component level, followed by correction procedures involving alignment to ensure text-image consistency, spatial cross-attention maps, and component quantities alignments.} 
  \label{fig:corrections}
  \vspace{-3mm}
\end{figure}

To further enhance the fine-grained details at the component level, GarmentAligner involves three component-level corrections (see \cref{fig:corrections}) into the training process of the diffusion model to provide feedback on the visual, spatial, and quantitative alignment for garment components.

\noindent \textbf{Visual alignment.} The CLIP Score \cite{hessel2022clipscore} is employed as a reward function to align the text and image of the components. Specifically, upon obtaining a generated image \(\hat X\), we identify the rough area of \(i\)-th component using the component mask \(m_i\) derived from the ground truth. Subsequently, we calculate the CLIP Score between the masked component area of \(\hat X\) and the corresponding text description \(t_{i}\). The visual correction loss is computed as the following equation:
\begin{equation}
    \label{visual align}
    {\cal L}_{\mathrm{visual}} =  {\sum_{i=1}^k \frac{1}{CLIPScore(m_i \odot {\hat X},  t_{i})}} ,
\end{equation}
where \(CLIPScore(x,t)\) denotes calculating the CLIP Score\cite{hessel2022clipscore} between image \(x\) and caption \(t\).

\noindent \textbf{Spatial alignment.} We extract spatial attention maps \(A_{i}=\{a_{i}^1, ..., a_{i}^l\}\) corresponding to the \(i\)-th component description \(t_{i}\) within the diffusion model. Subsequently, we leverage the component mask \(m_i\) obtained from the ground truth to guide \(a_{i}^j\) by computing the MSE loss. This strategy facilitates the model in learning the correspondence from a component position caption to the respective spatial location on the generated garment. The spatial correction loss can be expressed as:
\begin{equation}
    \label{spatial align}
    {\cal L}_{\mathrm{spatial}} = 
        \sum_{i=1}^k{
        \sum_{j=1}^l {
            || a_{i}^j - I_j(m_i) ||^2
            }
            % CLIPScore\left(m_i \odot {\hat X}, t_{c_i}\right)}
        },
\end{equation}
where \(I_j(\cdot)\) denotes the interpolation operation used to adjust the size of the component mask \(m_i\) to match that of \(a_{i}^j\).

\noindent \textbf{Quantitative alignment.} To achieve quantitative alignment, we utilize a component detector \cite{liu2023groundingdino} to determine recognition boxes for individual components, enabling the computation of component counts in the generated output.
Subsequently, these counts \(\hat Q = \{\hat q_1, ... , \hat q_k\}\) are compared with the ground truth component quantities \(Q = \{q_1, ... ,  q_k\}\) to rectify quantitative alignment at the component level, as depicted by the following equation:
\begin{equation}
    \label{count align}
    {\cal L}_{\mathrm{quantitative}} = 
        \sum_{i=1}^k {
            |q_i - \hat q_i|
            % CLIPScore\left(m_i \odot {\hat X}, t_{c_i}\right)
        }.
\end{equation}

The final loss function is determined by combining all correction losses and the retrieval-augmented contrasting loss:
\begin{equation}
    \label{count align}
    {\cal L} = 
        \omega_v \cdot {\cal L}_{\mathrm{visual}} +
        \omega_s \cdot {\cal L}_{\mathrm{spatial}} +
        \omega_q \cdot {\cal L}_{\mathrm{quantitative}} +
        \omega_r \cdot {\cal L}_{\mathrm{RACL}},
\end{equation}
where \(\omega_v, \omega_s, \omega_q, \omega_r\) represent the weights used to balance all losses.
\section{Experiments}
\subsection{Datasets}
Our experiments are conducted on the CM-Fashion dataset \cite{Zhang_2022armani}, which consists of 500,000 garment images at a resolution of 512×512, each accompanied by corresponding captions.
We employ the proposed automatic component extraction pipeline outlined in \cref{sec:acep} to extract component-level garment segmentation and count of components from the images. Subsequently, we enrich the captions with the extracted information. Consequently, we curate an augmented garment dataset derived from the CM-Fashion dataset \cite{Zhang_2022armani}, featuring optimized captions and component-level segmentation and quantity.

\subsection{Implementation Details}
GarmentAligner is implemented using PyTorch \cite{paszke2017pytorch} and trained on 8 Tesla V100 GPUs with a batch size of 32. The mask prediction network is adapted from ARMANI \cite{Zhang_2022armani}, renowned for its high segmentation accuracy, making it suitable for garment datasets. During training, we introduced the prediction of a mixture of noise and images as the prediction type, replacing the previous method of solely predicting noise. This enhancement is anticipated to elevate the quality of the generated outputs. We utilized SD v2.1\cite{rombach2022highresolution} as the base pre-trained model and employed a learning rate of \(1e-6\) for \(40\) epochs, requiring approximately 70 hours to complete the training process.

\subsection{Qualitative Results}
\begin{figure}[!t]
  \centering
  \includegraphics[scale=0.30]{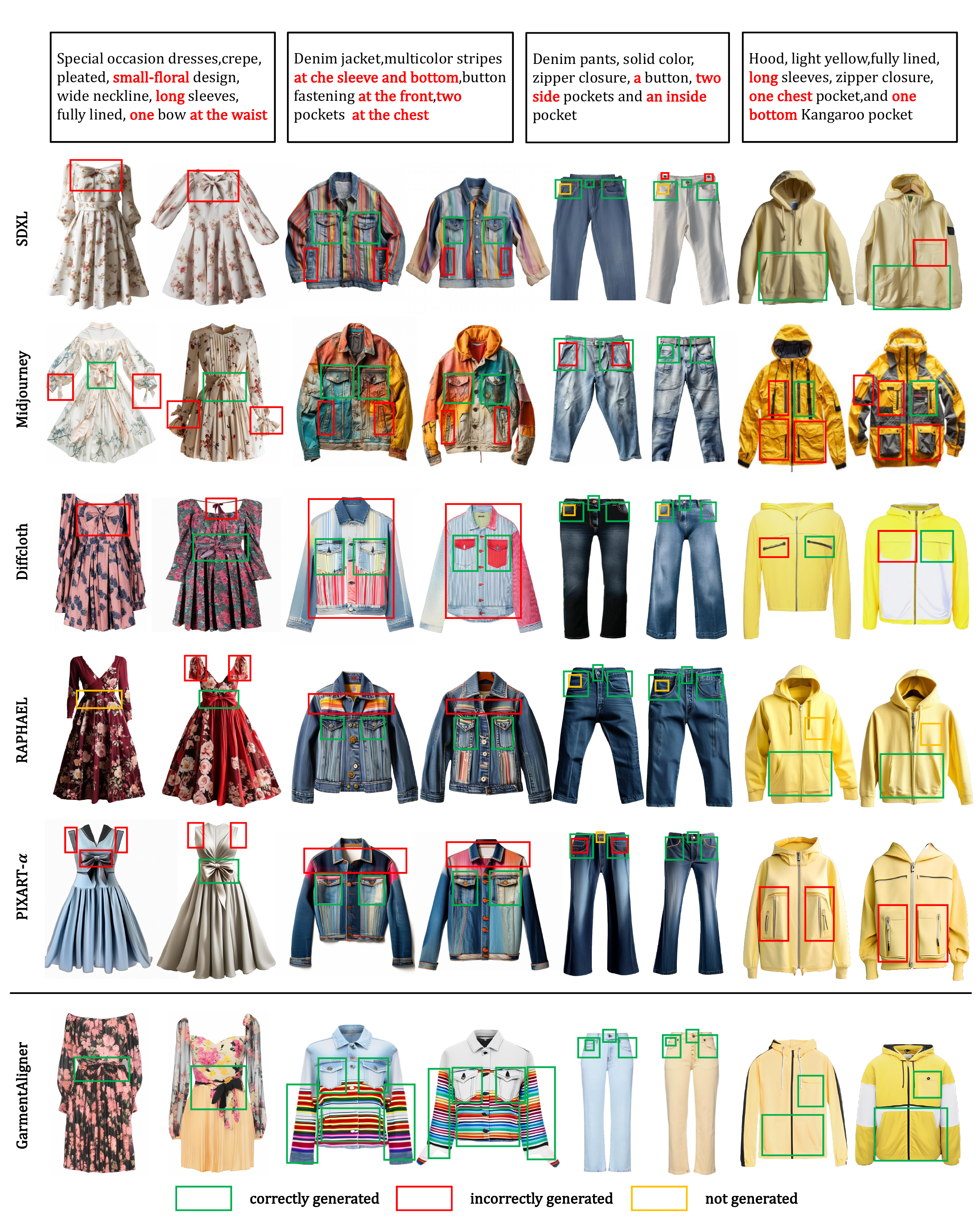}
  \caption{Visual comparison with baselines\cite{podell2023sdxl,zhang2023diffcloth,xue2024raphael,chen2023pixart}. Red boxes indicate incorrect generated areas, green boxes denote correct ones, and yellow boxes signify absent ones. Our approach demonstrates exceptional performance in capturing the texture, positioning, and quantity of garment components, resulting in the generation of realistic fashion images with precise fine-grained alignment.} 
  \label{fig:quality}
  \vspace{-3mm}
\end{figure}
We conduct a qualitative evaluation for GarmentAligner and state-of-the-art methods \cite{podell2023sdxl,zhang2023diffcloth,xue2024raphael,chen2023pixart}. The results depicted in \cref{fig:quality} demonstrate GarmentAligner's superiority in text-to-garment generation.  
While current state-of-the-art methods \cite{podell2023sdxl,zhang2023diffcloth,xue2024raphael,chen2023pixart} produce high-quality garment images aligned with the text description, they often overlook fine-grained information, particularly in quantity and position.
SDXL\cite{podell2023sdxl} and Midjourney prioritize detail and richness excessively, sometimes exceeding semantic relevance. Conversely, Diffcloth\cite{zhang2023diffcloth}, RAPHAEL\cite{xue2024raphael}, and PIXART-$\alpha$\cite{chen2023pixart} exhibit weaker semantic understanding of different garment components, hindering precise image generation.
% In contrast, SDXL\cite{podell2023sdxl} and Midjourney, though capable of generating high-quality images, tend to overly pursue detail and richness of image information, sometimes producing information beyond semantics. On the other hand, Diffcloth\cite{zhang2023diffcloth}, RAPHAEL\cite{xue2024raphael}, and PIXART-$\alpha$\cite{chen2023pixart} exhibit weaker semantic understanding of individual garment attributes, making it challenging to generate perfectly fitting images. 
Comparatively, GarmentAligner excels in generating realistic fashion images aligned with textual descriptions and exhibits precise fine-grained alignment. 
% GarmentAligner not only excels in generating realistic fashion images that align with textual descriptions but also boasts highly precise fine-grained alignment. GarmentAligner leverages the proposed Multi-level Correction to enable the model to perform real-time correction and feedback on various attribute information during training, enhancing the model's ability to learn fine-grained details. Additionally, through retrieval-enhanced techniques, the model can access more accurate component information for contrastive learning, resulting in images that almost perfectly match garment component attributes, particularly excelling in terms of quantity and spatial alignment.

\subsection{Quantitative Results}
\label{sec:quant}
\begin{table}[tb]
  \caption{The quantitative comparison results with baselines\cite{ramesh2021dalle,Zhang_2022armani,rombach2022highresolution,tang2023composable,feng2023StructureDiffusion,chefer2023attend,zhang2023diffcloth,podell2023sdxl}. Our approach achieves the best performance in comparison with numerous methods, excelling in metrics for image quality and consistency between text and image.}
  \label{tab:quantity}
  \centering
  \vspace{-2mm}
  \setlength{\tabcolsep}{2.6mm}{
  \begin{tabular}{lcccc}
    \toprule
    Method & FID$\downarrow$ & CLIPScore$\uparrow$ & AestheticScore$\uparrow$ & HPSv2$\uparrow$  \\
    \midrule
    DALL·E\cite{ramesh2021dalle} & 13.249 & 0.6423 &	4.8592 & 0.2137\\
    ARMANI\cite{Zhang_2022armani}	& 12.336 & 0.6988 &	5.3585 & 0.2237\\
    SDv1.5\cite{rombach2022highresolution}  & 9.368	 & 0.8911 & 5.2807 & 0.2419\\
    SDv2.1\cite{rombach2022highresolution}  & 9.157  & 0.8818 & 5.3881 & 0.2426\\
    ComposableDiffusion\cite{tang2023composable} & 9.499 & 0.8306 & 5.0984 &	0.2346\\
    StructureDiffusion\cite{feng2023StructureDiffusion}& 9.238	& 0.8459	& 5.2148	& 0.2398\\
    Attend-and-Excite\cite{chefer2023attend} & 9.351	& 0.8241	& 5.2452	& 0.2430\\
    DiffCloth\cite{zhang2023diffcloth}	& 9.201	& 0.8974	& 5.3957	& 0.2440\\
    SDXL\cite{podell2023sdxl} & 9.091	& 0.8756 	& 5.4299	 &0.2450\\
    GarmentAligner(ours) & \bf{8.735}	& \bf{0.9245}	& \bf{5.8776}& \bf{0.2648}\\
  \bottomrule
  \end{tabular}
  } 
  \vspace{-2mm}
\end{table}

\begin{figure}[!t]
  \centering
  \includegraphics[scale=0.40]{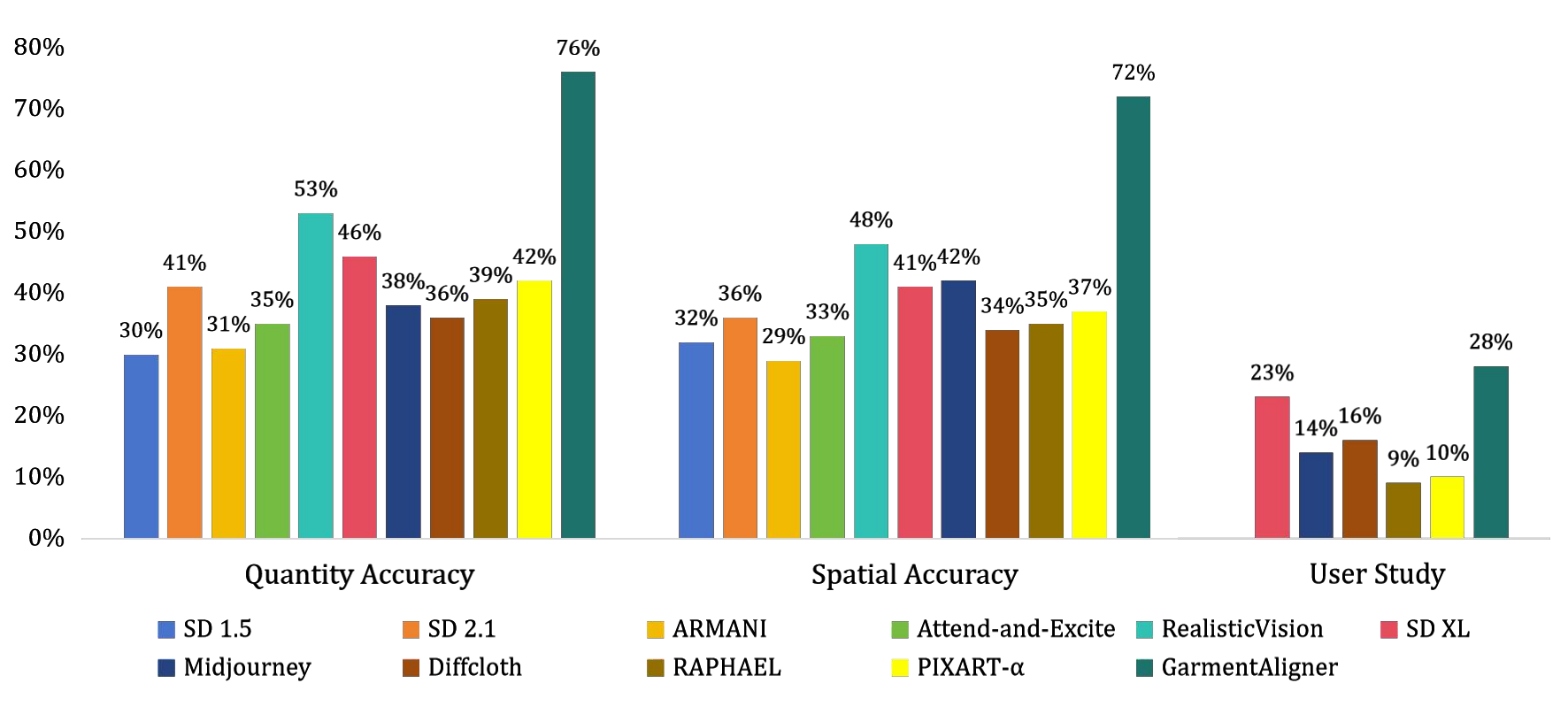}
  \caption{The component-level quantity and spatial accuracy of our method and baselines\cite{rombach2022highresolution, podell2023sdxl, zhang2023diffcloth, Zhang_2022armani, chefer2023attend, xue2024raphael, chen2023pixart} and user study result. Our approach demonstrates outstanding performance, significantly surpassing other methods in both quantity and spatial dimensions, and obtains user preferences.} 
  \label{fig:quanti}
\vspace{-3mm}

\end{figure}

For quantitative comparisons, we employ several metrics: 1) FID\cite{heusel2017fid} to assess the fidelity of the generated images, 2) Aesthetic Score to evaluate their aesthetic quality, 3) CLIP Score \cite{hessel2022clipscore} to measure the relevance between the given text and the generated images, and 4) HPS v2 \cite{wu2023hps} to gauge the degree of alignment with human preferences. The quantitative comparison results are presented in \cref{tab:quantity}, juxtaposed with those of DALL·E \cite{ramesh2021dalle}, ARMANI \cite{Zhang_2022armani}, ComposableDiffusion \cite{tang2023composable}, StructureDiffusion \cite{feng2023StructureDiffusion}, Attend-and-Excite \cite{chefer2023attend}, DiffCloth \cite{zhang2023diffcloth}, SD v1.5\cite{rombach2022highresolution}, SD v2.1\cite{rombach2022highresolution}, and SDXL \cite{podell2023sdxl}. Our proposed GarmentAligner achieves the lowest FID score and the highest scores in CLIPScore, Aesthetic score, and HPS v2 for garment synthesis, indicating superior performance across all metrics.

Additionally, we devised a component-level accuracy comparison to quantitatively assess the precision of component generation. Specifically, we employed our approach and baselines \cite{rombach2022highresolution, podell2023sdxl, zhang2023diffcloth, Zhang_2022armani, chefer2023attend, xue2024raphael, chen2023pixart} to generate images based on 1000 captions, and then calculate the quantity and spatial accuracy. To ensure the validity of the experiment, we generate 100 images for each caption. As depicted in \cref{fig:quanti}, our method outperforms others in accuracy by $20 \sim 45$\%, affirming the superiority of our approach in capturing fine-grained details of components.

% Besides, we devised a component-level accuracy comparison to quantitatively assess the precision of component generation. 
% Specifically, we generated multiple images for the same text and then marked the accuracy of the structural parts that matched the text description in these images. We selected 1000 texts with rich quantity and spatial information for testing, with each text generating 100 corresponding images. As shown in \cref{fig:quanti}, compared to other methods, our structural accuracy has improved by $20 \sim 45$\%, reaching the highest level.

\subsection{Ablation Study}
\begin{table}[tb]
  \caption{The quantitative comparison of different parts of the model,  where \([V]\) represents visual correction, \([S]\) represents spatial correction, \([C]\) represents quantitative correction, 
  % \([M]\) represents multi-level corrections 
  and \([R]\) represents retrieval-augmented contrastive learning.
  }
  \label{tab:abla}
  \centering
  \vspace{-2mm}
  \setlength{\tabcolsep}{2.4mm}{
  \begin{tabular}{lcccccc}
    \toprule
    Method & FID$\downarrow$ & CLIPScore$\uparrow$ & AestheticScore$\uparrow$ & HPSv2$\uparrow$  \\
    \midrule
    GarmentAligner$^{[V]}$&8.975	&0.9136	&5.4081	&0.2459\\
    GarmentAligner$^{[S]}$	&9.143	&0.8976	&5.4003	&0.2447\\
    GarmentAligner$^{[C]}$	&9.091	&0.8840	&5.3912	&0.2433\\
    GarmentAligner$^{[V+S+C]}$	&8.924	&0.9183	&5.4190	&0.2462\\
    GarmentAligner$^{[R]}$	&8.802	&0.8984	&5.7443	&0.2639\\
    GarmentAligner$^{[V+S+C+R]}$	&\bf{8.735}	&\bf{0.9245}	&\bf{5.8776}	&\bf{0.2648}\\
  \bottomrule
  \end{tabular}}
\end{table}

\begin{figure}[!t]
    \centering
    \begin{minipage}[t]{0.48\textwidth}
        \centering
        \includegraphics[width=6cm]{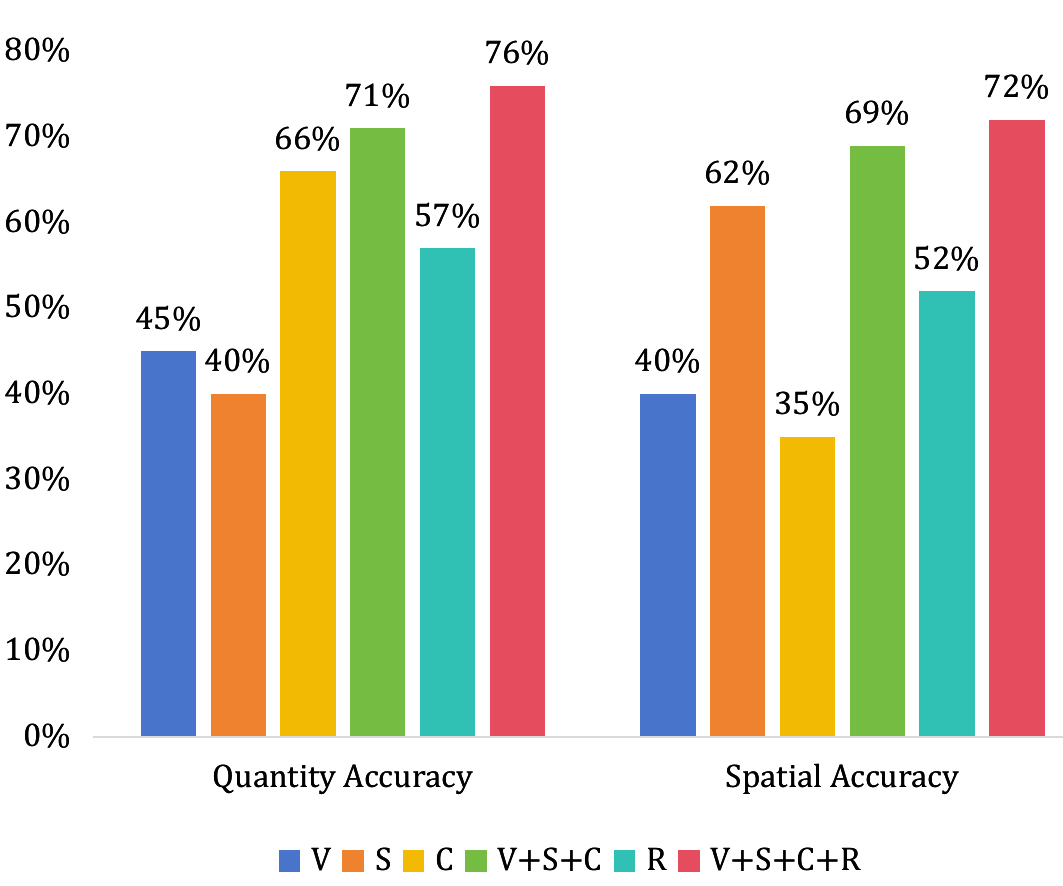}
        \caption{The component-level quantity and spatial accuracy of different variants from the ablation study. The multi-level corrections exhibit improvements for target attributes, while the combination of different parts contributes to a more effective enhancement.}
        \label{fig:abla1}
        \end{minipage}
        \hspace{1mm}
        \begin{minipage}[t]{0.48\textwidth}
        \centering
        \includegraphics[width=5.7cm]{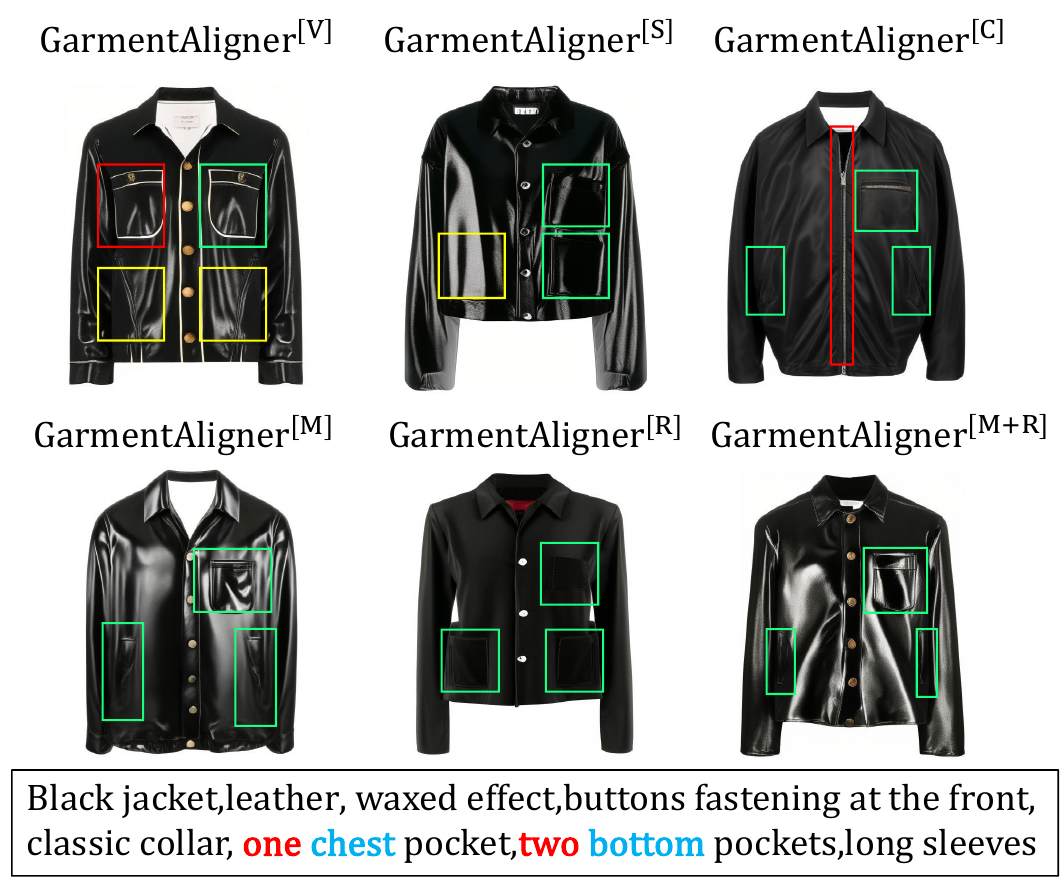}
        \caption{The visual comparison of different variants from the ablation study. \([M]\) denotes the combination of \([V+S+C]\), while the rest share equivalent definitions as illustrated in \cref{tab:abla}. The boxes maintain identical semantics with  \cref{fig:quality}.}
        \label{fig:abla2}
    \end{minipage}
    \vspace{-3mm}
\end{figure}

To validate the effectiveness of the proposed retrieval-augmented contrastive learning and the three corrections at the component level, we designed six variants of the proposed method and evaluated their performance based on the metric scores, accuracy and visual effects, as shown in \cref{fig:abla1}, \cref{fig:abla2} and \cref{tab:abla}. 
% We adopt the same metrics as \cref{sec:quant}. 

From \cref{tab:abla}, it is evident that retrieval-augmented contrastive learning makes the most significant contribution to the overall quality of generated garments, as evinced by its marked improvement across metrics such as FID\cite{heusel2017fid}, Aesthetic Score, and HPS v2\cite{wu2023hps}, which are indicative of image realism. On the other hand, multi-level corrections prove notably effective in enhancing text-image consistency while concurrently improving performance on all metrics. The accuracy outcomes illustrated in \cref{fig:abla1} demonstrate the significant role of multi-level corrections in enhancing both the quantity and spatial accuracy at the component level, thus validating the effectiveness of our proposed method design.
The combined application of these parts yields superior results, illustrating the complementary nature of the two pivotal components we propose. 
\cref{fig:abla2} illustrates the disparities among variant models in depicting detailed components of clothing, providing a more intuitive insight into the distinct roles played by different parts.

% According to \cref{tab:abla}, we enhance the consistency of generated images by introducing visual alignment, aligning text embeddings and image embeddings at the component-level. Subsequently, by introducing quantity and spatial alignment, we conduct real-time acquisition, inspection, correction, and feedback of the quantity and spatial information of the latent images generated during training, significantly improving the accuracy of quantity and spatial generation as shown in \cref{fig:abla1}. In summary, by introducing multi-level corrections and conducting component-level multi-feature alignment, we achieve better CLIPScore and structural accuracy. Additionally, we introduce Retrieval-augmented contrastive learning to increase the authenticity of generated images,and select comparison samples from multiple perspectives, ensuring that the generated images not only align with the text but also better adhere to aesthetic principles and human preferences.

% \begin{figure}[!t]
%   \centering
%   \includegraphics[scale=0.36]{IMG/inconsistency.pdf}
%   \caption{The visual comparison of results from different models of the ablation.} 
%   \label{fig:intro}
% \end{figure}

\subsection{User Study}

% \begin{figure}[!t]
%     \centering
%     \includegraphics[scale=0.36]{IMG/userstudy.pdf}
%     \caption{The human evaluation results of our method and baselines.Our approach has the highest human preference.} 
%     \label{fig:userstudy}
%     % \vspace{-3mm}
% \end{figure}

We executed a user study with 110 participants spanning a range of identities and age brackets to impartially evaluate our method against established baselines \cite{podell2023sdxl, zhang2023diffcloth, xue2024raphael, chen2023pixart}, with a focus on fine-grained fidelity and overall image quality.
We devise a questionnaire comprising 40 questions, each presenting shuffled outcomes generated from various methods associated with the same caption. Participants are instructed to select the image they perceive to be the most authentic and consistent with the provided textual description. The results, shown in \cref{fig:quanti}, reveal that our method was the preferred choice for the majority of participants, garnering selections exceeding 28\%.

\section{Conclusion}
In this study, we introduce GarmentAligner, a text-to-garment diffusion model aimed at rectifying fine-grained semantic misalignment issues inherent in garment generation.
By combining retrieval-augmented contrastive learning with multi-level corrections, GarmentAligner effectively aligns the visual semantics, spatial positioning, and quantity of the generated garment components with provided captions. Furthermore, we devise an automatic component extraction pipeline to extract spatial and quantitative information about garment components from images, which can be applied to any garment dataset, thus facilitating the advancement of high-quality garment generation.
Our experiments demonstrate that GarmentAligner produces superior garment images with improved semantic alignment compared to existing methods.

\noindent \textbf{Social Impacts and Limitations.} While producing high-quality garments, our method still encounters certain limitations. 
% Our training data is derived from the proposed extraction pipeline, which relies on the accuracy of semantic segmentation and detection models. Inevitable errors will occur when dealing with large-scale datasets. Furthermore, text-driven generative approaches are susceptible to biases from pre-trained models, potentially yielding results that are insufficiently secure and user-friendly, thus posing potential risks of misuse.
The training data utilized in our approach is generated through the proposed extraction pipeline, which heavily relies on the accuracy of semantic segmentation and detection models. Inevitably, errors may arise, particularly when dealing with large-scale datasets. Moreover, the biases inherited from pre-trained models potentially lead to outputs that lack robustness and user-friendliness. Future measures are necessary to address these ethical concerns through bias reduction and thorough scrutiny.

\section*{Acknowledgements}
% Please insert your acknowledgments here.
This work was supported in part by National Science and Technology Major Project (2020AAA0109704), National Science and Technology Ministry Youth Talent Funding No. 2022WRQB002, National Natural Science Foundation of China under Grant No. 62372482 and No. 61936002, Guangdong Outstanding Youth Fund (Grant No. 2021B1515020061), Shenzhen Science and Technology Program (Grant No. GJHZ20220913142600001), Mobility Grant Award under Grant No. M-0461, Nansha Key RD Program under Grant No.2022ZD014.
\bibliographystyle{splncs04}
\bibliography{main}
\end{document}